\documentclass[journal]{IEEEtran}
\usepackage{xcolor}
\usepackage[utf8]{inputenc}
\usepackage{subfigure}

\usepackage[final]{graphicx}
\graphicspath{{images/}}

\ifCLASSINFOpdf
\else
   \usepackage[dvips]{graphicx}
\fi
\usepackage{amsmath}
\usepackage{algorithmic}
\usepackage{multicol}

\ifCLASSOPTIONcompsoc
 \usepackage[caption=false,font=normalsize,labelfont=sf,textfont=sf]{subfig}
\else
 \usepackage[caption=false,font=footnotesize]{subfig}
\fi

\usepackage{stfloats}
\usepackage{url}
\author{Igor~Markov,
        Sergey~Nesteruk,
        Andrey~Kuznetsov,
        and Denis~Dimitrov
\thanks{ I. Markov is with AIRI }%
\thanks{ S. Nesteruk, A. Kuznetsov, and D. Dimitrov are with Sber AI }
}% <-this % stops a space

\title{RusTitW: Russian Language Text Dataset for Visual Text in-the-Wild Recognition}

\begin{document}
\markboth{March~2023}%
{Shell \MakeLowercase{\textit{et al.}}: Bare Demo of IEEEtran.cls for IEEE Journals}



\maketitle

\begin{abstract}
Information surrounds people in modern life. 
Text is a very efficient type of information that people use for communication for centuries. 
However, automated text-in-the-wild recognition remains a challenging problem. 
The major limitation for a DL system is the lack of training data.
For the competitive performance, training set must contain many samples that replicate the real-world cases.
While there are many high-quality datasets for English text recognition; there are no available datasets for Russian language.
In this paper, we present a large-scale human-labeled dataset for Russian text recognition in-the-wild. 
We also publish a synthetic dataset and code to reproduce the generation process\footnote{Code and data is avaliable at \url{github.com/markovivl/SynthText}}. 
\end{abstract}

% \fi
\IEEEpeerreviewmaketitle

\section{Introduction}
\IEEEPARstart{I}{nformation} surrounds people everywhere in modern life. 
Text is a very efficient type of information that people have used for communication for centuries. 
AI-based technologies help us to automate data processing. 
Data science community developed models that reach super-human accuracy on the variety of tasks. 
However, automated text-in-the-wild recognition remains a challenging problem. 


The complications originate from the complex nature of the real-world scenes. 
Text be hard to see besides other dominant objects on an image. 
The background can have too high color variation. 
Text itself can be curved, bent and assume different forms. 
It can be visually noisy due to imperfect lighting conditions, overlappings, etc.

The major limitation for a DL system is the lack of training data. 
For the competitive performance, training set must contain many samples that replicate the real-world cases.
While there are many high-quality datasets for English text recognition, there are no available datasets for Russian language. 
In this paper, we present a large-scale human-labeled dataset for Russian text recognition in-the-wild. 
We also publish a synthetic dataset and code to reproduce the generation process. 

There are many applications for text recognition in images~\cite{long2021scene} and in video~\cite{yin2016text}.
They include: water meter number reading~\cite{8606091}, content-based image retrieval~\cite{schroth2011exploiting}, robot navigation~\cite{schulz2015robot}, transportation~\cite{chowdhury2013extracting}, interactive translation~\cite{hsueh2011interactive}, vision-impaired people assistance~\cite{khan2020ai}.
The code can be easily adjusted for a specific problem. 

The rest of this paper is organized as follows.
In Section~\ref{sec:related_work} we overview the existing text datasets.
In Section~\ref{sec:titw_dataset} we describe human-annotated Russian text recognition dataset.
In Section~\ref{sec:synth_dataset} we describe a tool for synthetic Russian text dataset generation, and describe the created dataset.
% In Section~\ref{sec:experiments} we conduct experiments with both datasets.

\section{Related Work}
\label{sec:related_work}

\subsubsection{ Natural Scene Text Datasets }

Datasets form the basis of AI development.
It is essential to have a relevant dataset to train an accurate model.
There are publicly available datasets for text recognition.

Street View House Numbers~(SVHN) dataset~\cite{netzer2011reading} is obtained from house numbers in Google Street View images.
It has 10 digits classes and 100K annotated digits in total.
Char dataset~\cite{deCampos09} has over 7K characters obtained from natural images.

SCUT-CTW1500 dataset~\cite{yuliang2017detecting} consists of 1.5K manually picked text images from the Internet.
Total-Text dataset~\cite{ch2017total} contains over 1.5K images with text.
The samples in this dataset can be curved and cover various text orientations.

DOST dataset~\cite{he2018end} contains 30K frames for text recognition in video sequences.

COCO-Text~\cite{gomez2017icdar2017} is based on MS~COCO dataset and contains 173,589 labeled text regions in over 63K images.
TextOCR dataset~\cite{singh2021textocr} 900K annotated words over 28K images. 
OpenTextImages~\cite{krylov2021open} is a large text recognition dataset with multiple tasks.
It consists of more than 207K images and 2.5M instances.
Uber-Text dataset~\cite{zhang2017uber} has 110K street view images with text.

HierText~\cite{long2022towards} is the dataset featuring hierarchical annotations of text in natural images.
The goal of the dataset is to detect text as masks and further group them into clusters.

% Street View Text (SVT)

There are datasets that cover not only English texts.
ICDAR~\cite{karatzas2013icdar} is a Robust Reading Competition launched in 2003.
It started with 509 natural scene images with text, and significantly progressed over the years.
Now it is a multi-lingual benchmark for text recognition~(MLT-19)~\cite{nayef2019icdar2019}.
It has more that 20K images with annotations for 4 text-related tasks.
Chinese Street View Text~\cite{sun2019chinese} is a large-scale dataset with partial supervision. 
It contains 430K images.
30K images are fully annotated.
However, these datasets do not have Russian text or any other language with Cyrillic script.

\subsubsection{ Synthetic Text Datasets }

When there are no relevant training datasets, one can either collect a new dataset~\cite{XtremeAugment} or generate a synthetic one~\cite{dandekar2018comparative}.
Dataset collection and annotation is very time-consuming and expensive.
For some problems it is easier to generate a dataset that is suitable for model training.
There are many synthetic datasets for computer vision tasks. 
They cover both indoor~\cite{roberts2021hypersim} and outdoor environment~\cite{ros2016synthia}.

One can also find synthetic datasets for the text recognition task.
UnrealText dataset~\cite{long2020unrealtext} has 600K synthetic images with text embedded into modelled scene surfaces.
There are more than 12M cropped text instances in this dataset.

Synth90k~\cite{jaderberg2014synthetic} is a synthetic word dataset. 
It has 9M word-level crops. 
They apply basic image augmentations such as projective distortion, simple blending and noise.

SynthText dataset~\cite{synthtext} has over 800K images, and the total of 6M words.
This is the largest text dataset and a popular tool for dataset generation.
However, it requires either pre-computed depth and segmentation maps or switching between multiple programming languages, namely Python and Matlab, which is proprietary.

The described synthetic datasets do not support Russian language.
To the best of our knowledge, there are only several Russian-language datasets, any they are for handwritten text only.
The dataset of Peter the Great’s manuscripts contains 9694 images~\cite{potanin2021digital}. 
Russian and Kazakh handwritten text dataset HKR has 63K short strings by 200 writers~\cite{nurseitov2021handwritten}.

% {\color{red} No Russian datasets }

% {\color{red} No end-to-end generation tools }

% ~\cite{liu2020abcnet}

%\IEEEpubidadjcol

% \section{  Datasets }

\section{ Russian Text in the Wild Dataset }
\label{sec:titw_dataset}

\begin{figure}[!t]
\centering
\setlength\fboxsep{0pt}
\setlength\fboxrule{0.25pt}
\fbox{\includegraphics[width=0.475\textwidth]{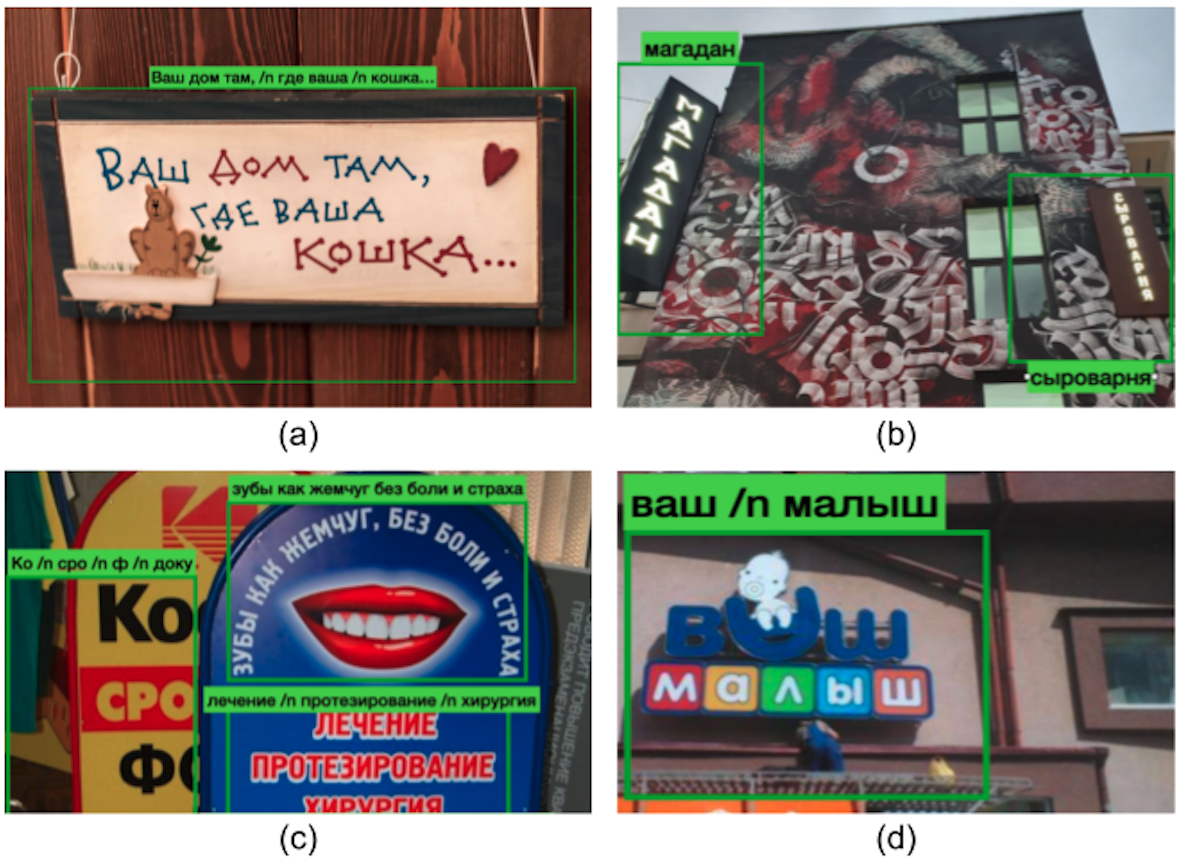}}
\caption{Example of an annotated image.}
\label{fig:titw_examples}
\end{figure}

In this work, we share both a real-world text in-the-wild dataset with human annotations and a synthetic dataset.
The real-world dataset contains over 13K images accompanied with paragraph-level annotations.
Each paragraph annotation has a bounding box and single text string in~Fig.{\ref{fig:titw_examples}}.
Bounding boxes are not tight to include text background. 
Text annotations include digits and common punctuation (.,?!:;-). 
% Letter "ё"  is distinguished from letter "е".
Other special characters are ignored. 
Text is case-sensitive. 
We annotate both Cyrillic and Latin letters. 
Vertical text has separate bounding box for each column. 
Very small and unclear text is ignored.

% \begin{figure*}
% \centering
% \includegraphics[width=0.9\textwidth]{images/titw_samples.png}
% \caption{Examples of images with annotations.}
% \label{fig:titw_examples}
% \end{figure*}

% \begin{table}[!t]
% % increase table row spacing, adjust to taste
% \renewcommand{\arraystretch}{1.3}
% \caption{Russian Tet in the Wild Dataset Statistics}
% \label{tab:titw_dataset_stats}
% \centering

% \begin{tabular}{|c||c|c|c|}
% \hline
% Subset & Training & Test & Joint \\
% \hline
% \#images & 9442 & 14673 & 24115 \\
% \hline

% \#boxes & 25705 & 38188 & 63893 \\
% \hline

% \#lines & 43815 & 58484 & 102299 \\
% \hline

% \#words & 98942 & 106022 & 204964 \\
% \hline

% \#unique words & 37585 & 37267 & 65056 \\
% \hline

% \#unique words (case sensitive) & 31369 & 30706 & 52813 \\
% \hline

% \#unique words without numbers & 25998 & 26197 & 43545 \\
% \hline
 
% \end{tabular}
% \end{table}

\begin{table}[!t]
\renewcommand{\arraystretch}{1.3}
\caption{Russian Text in-the-Wild Dataset Statistics}
\label{tab:titw_dataset_stats}
\centering

\begin{tabular}{|c||c|c|c|}
\hline
Subset & Training & Test & Joint \\
\hline
\#images & 10000 & 3795 & 13795 \\
\hline

\#boxes & 27645 & 10739 & 38384 \\
\hline

\#boxes with Russian text & 8155 & 3098 & 11253 \\
\hline

\#boxes with English text & 3483 & 1311 & 4794 \\
\hline

\#boxes with digits & 3441 & 1316 & 4757 \\
\hline

\#boxes with punctuation & 4217 & 1567 & 5784 \\
\hline

\#lines & 46479 & 17867 & 64346 \\
\hline

\#words & 96810 & 36324 & 133134 \\
\hline

\#unique words (case sensitive) & 33504 & 15739 & 42387 \\
\hline

% \#unique words  & 000 & 000 & 000 \\
% \hline

\#unique words without numbers & 26804 & 13157 & 33445 \\
\hline
 
\end{tabular}
\end{table}

In Table~\ref{tab:titw_dataset_stats} one can find the statistics of the dataset.

% \section{ Method }
\section{ Synthetic Russian Text Dataset }
\label{sec:synth_dataset}

% {\color{red} Generation pipeline }
\subsection{ Generation pipeline }

Our pipeline is based on the SynthText dataset generation pipeline~\cite{synthtext}.
We modified it for higher flexibility and better visual realism.

To exclude missing annotations, we must consider that original images can already contain text.
To detect such cases, we use the CRAFT model~\cite{craft}. It is a character-level text detection model.
With text bounding boxes we can either filter an image out or blur the detected text parts before adding new text.

To find an appropriate place for text insertion, we assume that usually text is located in uniform regions.
We estimate uniform regions using boundary detection with the~(COB) model~\cite{cob}.
It produces hierarchical boundaries, which allows us to choose a scale of the selected regions.

Text looks more realistic when it follows the shape of the objects in the original scene.
For scene geometry estimation, we use the monocular depth estimation model~(MiDaS)~\cite{midas}.
It is robust for various datasets, and gives a relative depth for every pixel.

As in the original pipeline in~\cite{synthtext}, for each image we use the semantic segmentation map to determine the most suitable regions.
We obtain them with model in~\cite{cob}.
After the candidate regions are determined, text is sampled from the text dataset, details of which we present in the next subsection.
During sampling, text size and font are randomized. The candidate text render is generated at this stage.
After that, for each region the size text image is compared to the size of the region.
If the text image fits in the region, the reverse homography transformation is applied.
During calculation of homography, depth map is taken into account.
As a result, it allows text to blend more naturally into the image perspective-wise.
After that, text image is blended into the original image.
As in~\cite{synthtext}, we also use Poisson blending during this stage. Because of it, text repeats the texture and color features of original image region.
As a result, the text on the output image looks smoother.

\subsection{ Dataset Diversity }
The source of background images in our dataset is LAION dataset~\cite{laion}.
We severely filter it, and blur people's faces and existing text. 
One can find the statistics of the synthetic dataset in Table~\ref{tab:titw_dataset_stats}.
We use 96 Russian text fonts across 40 font families. 
Our text dataset is the dictionary of most frequent Russian words, which we filter from obscenities. We generate random number sequences of various size, add synthetic phone numbers and most frequent Russian surnames. 

Our framework allows to regulate sampling probabilities of each word.
However, in a final pipeline, we assigned the same probability for each word to obtain uniform sampling. The final statistics in our synthetic dataset is shown in~Table~\ref{tab:synth_dataset_stats}.

\begin{table}[!t]
\renewcommand{\arraystretch}{1.3}
\caption{Russian Synthetic Text in-the-Wild Dataset Statistics}
\label{tab:synth_dataset_stats}
\centering

\begin{tabular}{|c||c|c|c|}
\hline
Subset & Training & Test & Joint \\
\hline
\#images & 852092 & 48609 & 900701 \\
\hline

\#boxes & 1510481 & 86776 & 1597257 \\
\hline

\#boxes with Russian text & 1403096 & 80480 & 1483576 \\
\hline

\#boxes with English text & 95561 & 5368 & 100929 \\
\hline

\#boxes with digits & 191461 & 10980 & 202441 \\
\hline

\#boxes with punctuation & 338392 & 19285 & 357677 \\
\hline

\#lines & 2559573 & 146641 & 2706214 \\
\hline

\#words & 2958790 & 168022 & 3126812 \\
\hline

\#unique words (case sensitive) & 68439 & 52045 & 68448 \\
\hline

\#unique words without numbers & 64084 & 48725 & 64093 \\
\hline

\end{tabular}
\end{table}

% {\color{red} Text dataset: Russian + English }
% For dataset of Russian words we chose as a basis the 
% {\color{red} Fonts }
% 113

% {\color{red} Transformations and Augmentations }

% {\color{red} Orientation }

% {\color{red} Colors, borders }

% {\color{red} Generated data format (string / boxes / masks / ...) }

\begin{figure}
\setlength\fboxsep{0pt}
\setlength\fboxrule{0.25pt}
\centering
\fbox{\includegraphics[width=0.4\textwidth, height=5.5cm]{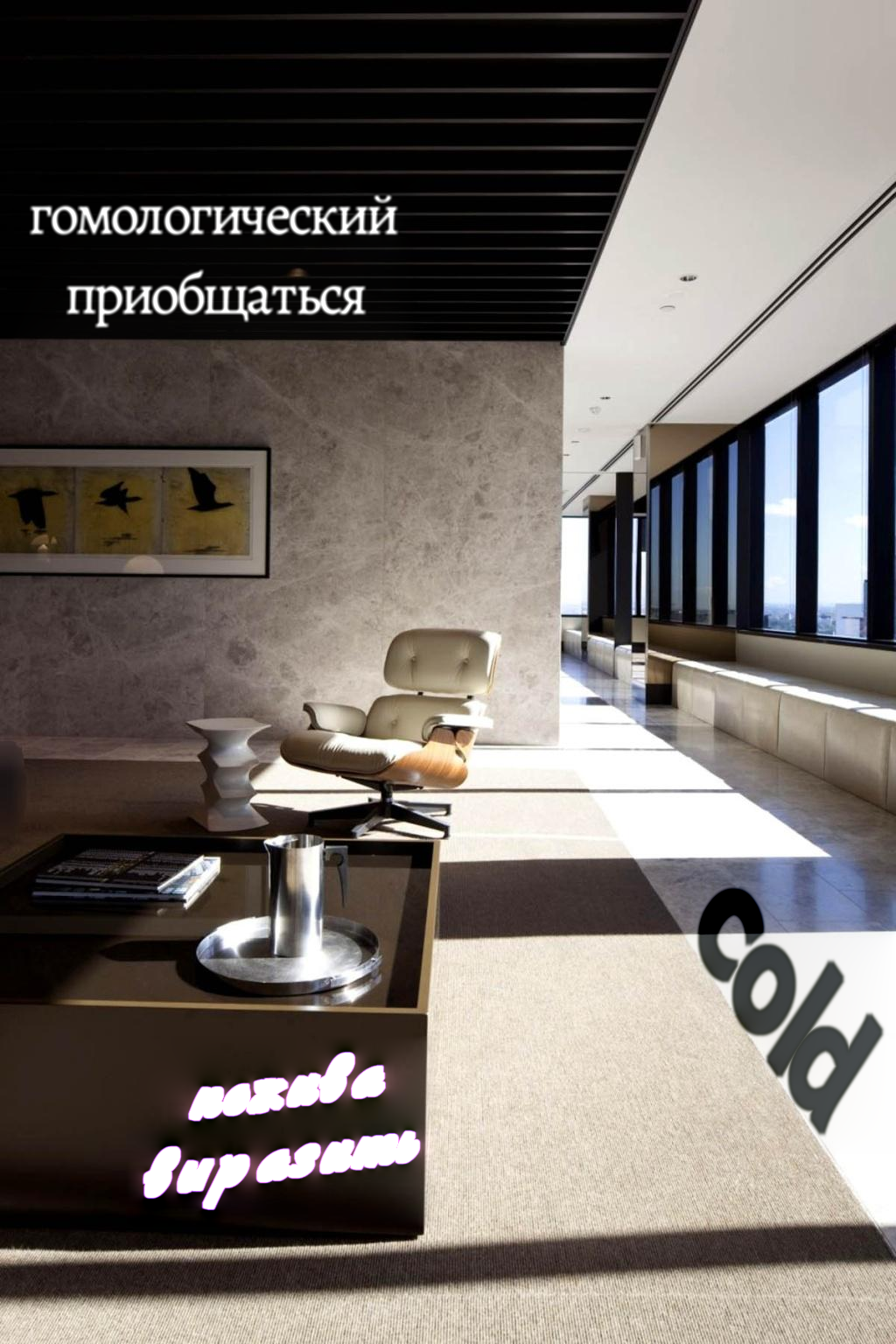}}
\caption{Example of synthetic text image which accounts for segmentation map and depth map.}
\label{fig:pics1}
\end{figure}

\begin{figure}
\setlength\fboxsep{0pt}
\setlength\fboxrule{0.25pt}
\centering
\fbox{\includegraphics[width=0.4\textwidth, height=6cm]{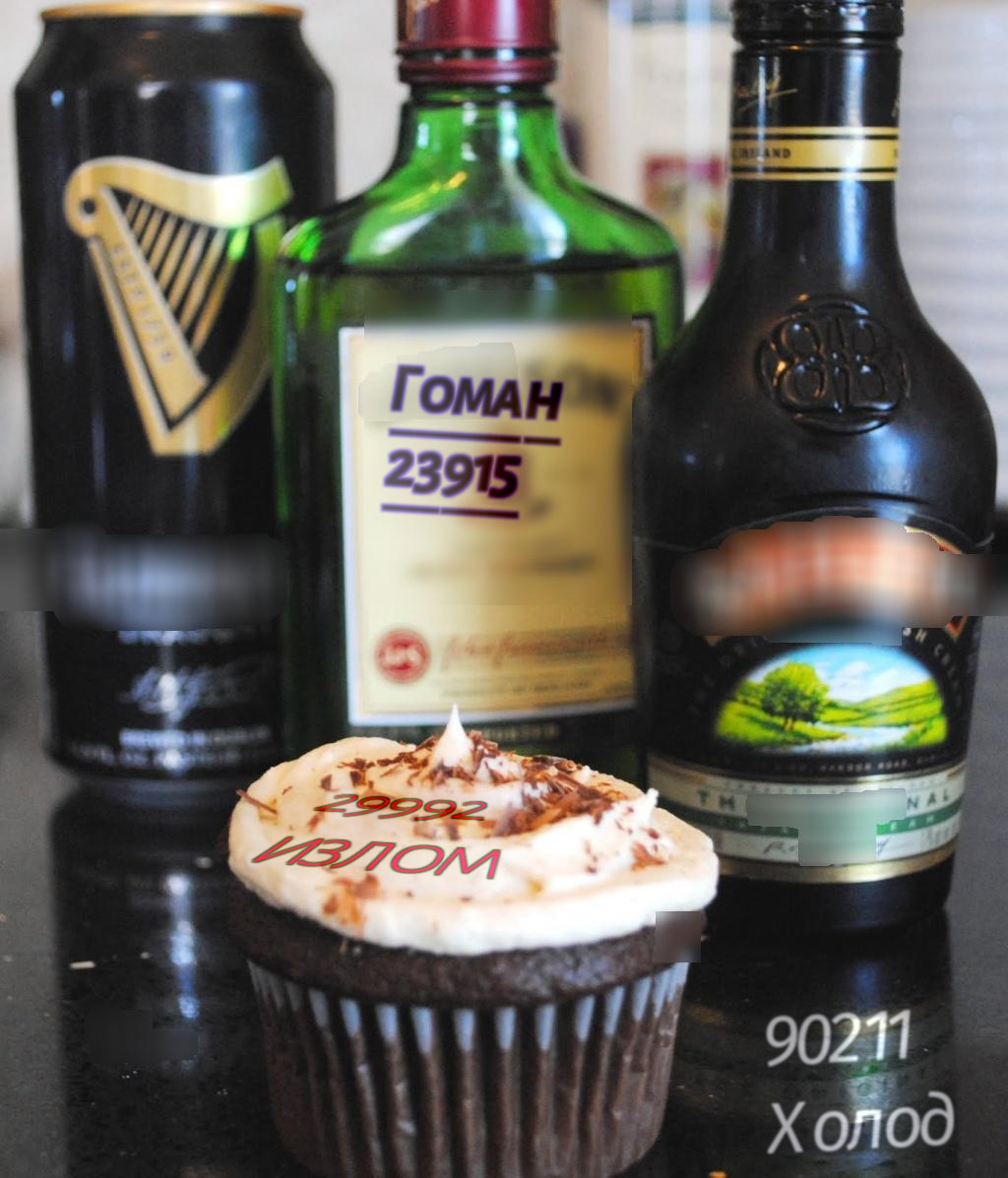}}
\caption{{Example of synthetic text image conditioned on segmentation and depth map with blurred out original text .}}
\label{fig:pics2}
\end{figure}

\subsection{ Generated Samples }
\begin{figure}
\setlength\fboxsep{0pt}
\setlength\fboxrule{0.25pt}
\centering
\fbox{\includegraphics[width=0.4\textwidth, height=6cm]{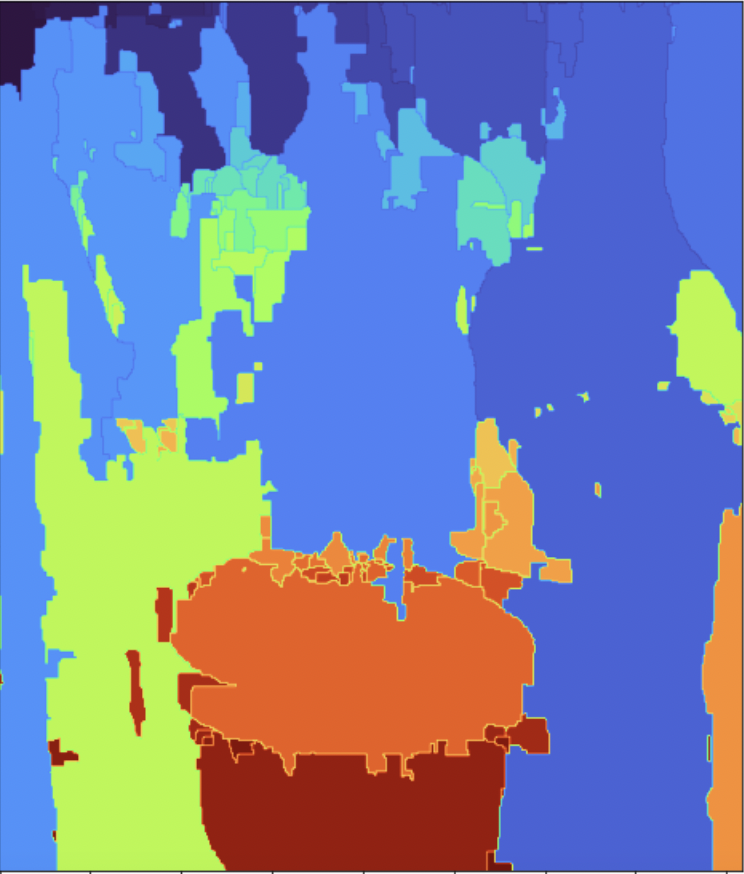}}
\caption{Example of segmentation map used for synthesis of Fig.~\ref{fig:pics2}.}
\label{fig:info1}
\end{figure}

Segmentation maps add spatial information to the algorithm~(Fig.~\ref{fig:info1}), which allows for a more meaningful text synthesis on the images. On~Fig.~\ref{fig:pics1} one can see how on the synthetic text on the top occupies spatially motivated regions.

\begin{figure}
\setlength\fboxsep{0pt}
\setlength\fboxrule{0.25pt}
\centering
\fbox{\includegraphics[width=0.4\textwidth, height=6cm]{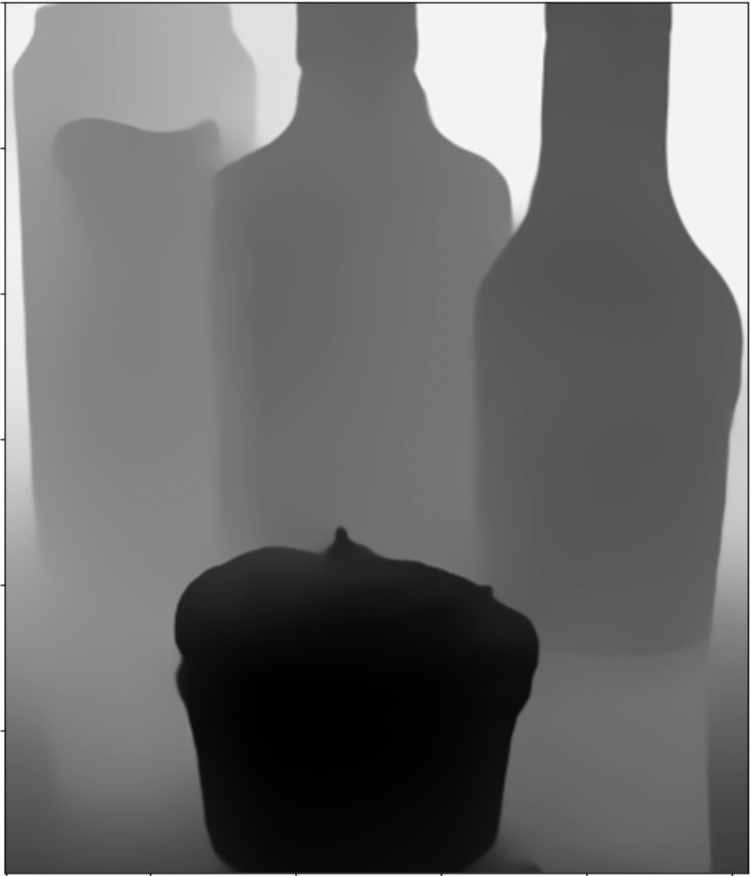}}
\caption{Example of depth map used for synthesis of Fig.\ref{fig:pics2}.}
\label{fig:info2}
\end{figure}

On~Fig.~\ref{fig:pics2} there is a lot of text on the original image, which normally would conflict with the synthetic data.
Naturally, such image would just be discarded. However, original text on such images already occupies very convenient regions. Such regions, by definition, are good areas for synthetic text.
By blurring out the original text, we generate synthetic one on an already suitable surface.
In addition to this, this image is the example of how the depth features (Fig.~\ref{fig:info2}) of the region are taken into account.

\begin{figure}
\setlength\fboxsep{0pt}
\setlength\fboxrule{0.25pt}
\centering
\fbox{\includegraphics[width=0.4\textwidth, height=6cm]{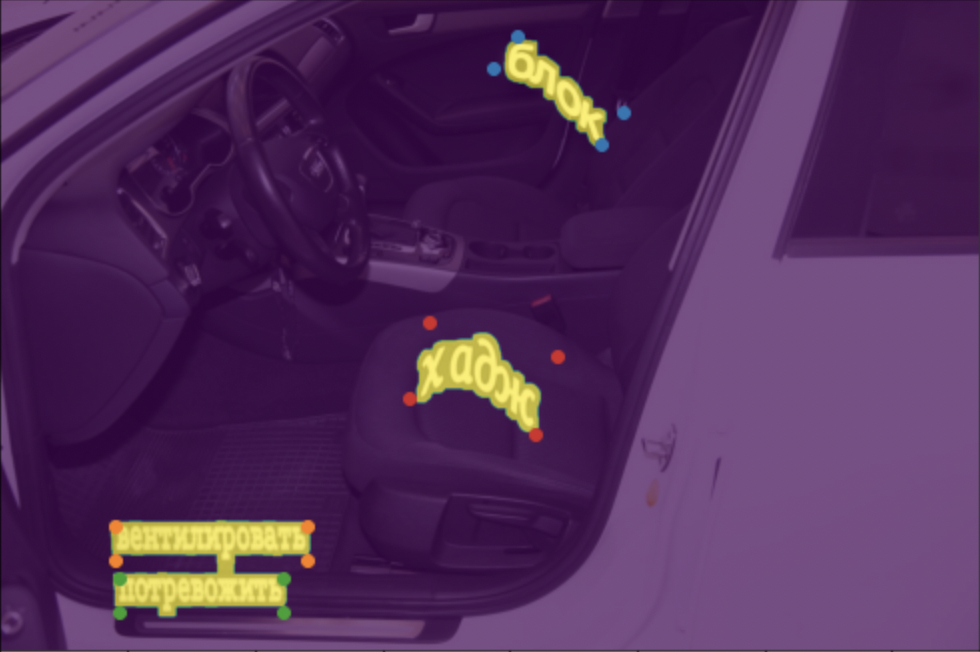}}
\caption{Example of an image with trigonometrically transformed synthetic text with paragraph-wise bounding boxes and segmentation labelling.}
\label{fig:label1}
\end{figure}

During generation we added extra trigonometric transformations (Fig.~\ref{fig:label1}) to the text, making the text fit to a randomized section of sine curve, additionally varying the resulting data.

Because of the synthetic approach, we are able to provide various range of labels for the text.
They can be used for out-of-the-box training for wide range of models, which rely on different labels.
As in the original pipeline by~\cite{synthtext}, we also provide paragraph-wise and character-wise bounding boxes, albeit we adapt them to conform with the aforementioned trigonometrical transformations.
In addition to this, we provide segmentation maps which highlight the text on the image~(Fig.~\ref{fig:label1}).

% \begin{figure}
% \setlength\fboxsep{0pt}
% \setlength\fboxrule{0.25pt}
% \centering
% \fbox{\includegraphics[width=0.4\textwidth, height=6cm]{images/label_ex_1.png}}
% \caption{Example of paragraph-wise bounding boxes.}
% \label{fig:label1}
% \end{figure}

% \begin{figure}
% \setlength\fboxsep{0pt}
% \setlength\fboxrule{0.25pt}
% \centering
% \fbox{\includegraphics[width=0.4\textwidth, height=6cm]{images/charbb_1.png}}
% \caption{Example of character-wise bounding boxes.}
% \label{fig:charbb1}
% \end{figure}

% \begin{figure}
% \setlength\fboxsep{0pt}
% \setlength\fboxrule{0.25pt}
% \centering
% \fbox{\includegraphics[width=0.4\textwidth, height=6cm]{images/segout_1.png}}
% \caption{Example of a segmentation map label.}
% \label{fig:segout1}
% \end{figure}

% {\color{red} Examples and Comparison }

% {\color{red} Highlight features }

% \section{ Experiments }
% \label{sec:experiments}

% Train on synthetic data, test on natural

% {\color{red} Datasets: }

% {\color{red} Baseline (only natural text images) }

% {\color{red} Synthetic text on uniform background }

% {\color{red} Synthetic text on natural images }

% {\color{red} Realistic synthetic text on natural images }

% {\color{red} Realistic synthetic text on natural images + heuristics(filter bad samples / old depth vs. MiDaS) }

% {\color{red} Mix of the previous }

% ...

% {\color{red} Models: }

% {\color{red} Mango }

% {\color{red} Transformer }

% {\color{red} Bezier curve models }

\section{Conclusion}
We created the first large-scale Russian language dataset of text in-the-wild, using both 14k of real and more than 900k synthetically generated images.
This dataset can be used to finetune and improve the existing models and be a benchmark for future research. We also release the code, which can be used to reproduce our results and generate new datasets for all kinds of computer vision tasks requiring scene-text samples.

\ifCLASSOPTIONcaptionsoff
  \newpage
\fi



%\bibliographystyle{IEEEtran}



% \end{thebibliography}

\bibliographystyle{IEEEtran}
\bibliography{main.bib}

% \appendices
% %\section{P}
% \label{appendix:images}


% \end{IEEEbiography}


%\newpage



%\vfill


\end{document}